\documentclass[journal,]{IEEEtran}
\usepackage{cite}
\usepackage[utf8]{inputenc} 
\usepackage[T1]{fontenc}    
\usepackage{url}            
\usepackage{booktabs}       
\usepackage{amsfonts}       
\usepackage{nicefrac}       
\usepackage{microtype}      

\usepackage{graphicx}
\usepackage{comment}
\usepackage{amsmath,amssymb} 
\usepackage{color}
\usepackage{algorithm}
\usepackage{algpseudocode} 
\usepackage{amsmath,graphicx,setspace}
\usepackage{multirow, array, amssymb,booktabs,makecell,xspace,bm}
\usepackage{amssymb}
\usepackage{pifont}
\usepackage{amsfonts}
\usepackage{arydshln}
\usepackage{makecell}
\usepackage{breqn}
\usepackage{amsmath,bm}
\usepackage{bbm}

\newcommand{\etal}{\textit{et al.}}
\newcommand{\ie}{\textit{i.e.,}}
\newcommand{\eg}{\textit{e.g.,}}
\newcommand{\etc}{\textit{etc.}}

\usepackage[colorlinks = true,
            linkcolor = black,
            urlcolor  = black,
            citecolor = black,
            anchorcolor = black]{hyperref}
\newcommand{\MYhref}[3][blue]{\href{#2}{\color{#1}{#3}}}

\ifCLASSINFOpdf
\else
\fi

\begin{document}
%
\title{Integrating Language-Derived Appearance Elements with Visual Cues in Pedestrian Detection}

%
%
%

\author{Sungjune Park$^{\dagger}$, Hyunjun Kim$^{\dagger}$, and Yong Man Ro,~\IEEEmembership{Senior Member,~IEEE}%
\thanks{\textbf{This manuscript has been accepted for the publication in \textit{IEEE Transactions on Circuits and Systems for Video Technology (2024)}.} \\ S. Park, H. Kim, and Y. M. Ro are with Image and Video Systems Lab., School of Electrical Engineering, Korea Advanced Institute of Science and Technology (KAIST), 291 Daehak-ro, Yuseong-gu, Daejeon, 34141, Republic of Korea (e-mail: sungjune-p@kaist.ac.kr; kimhj709@kaist.ac.kr; ymro@kaist.ac.kr). \\
Corresponding author: Y. M. Ro. (ymro@kaist.ac.kr) \\
$^{\dagger}$Both authors contributed equally to this manuscript.}}

\maketitle

\begin{abstract}
Large language models (LLMs) have shown their capabilities in understanding contextual and semantic information regarding knowledge of instance appearances. In this paper, we introduce a novel approach to utilize the strengths of LLMs in understanding contextual appearance variations and to leverage this knowledge into a vision model (here, pedestrian detection). While pedestrian detection is considered one of the crucial tasks directly related to our safety (\textit{e.g.,} intelligent driving systems), it is challenging because of varying appearances and poses in diverse scenes. Therefore, we propose to formulate language-derived appearance elements and incorporate them with visual cues in pedestrian detection. To this end, we establish a description corpus that includes numerous narratives describing various appearances of pedestrians and other instances. By feeding them through an LLM, we extract appearance knowledge sets that contain the representations of appearance variations. Subsequently, we perform a task-prompting process to obtain appearance elements which are guided representative appearance knowledge relevant to a downstream pedestrian detection task. The obtained knowledge elements are adaptable to various detection frameworks, so that we can provide plentiful appearance information by integrating the language-derived appearance elements with visual cues within a detector. Through comprehensive experiments with various pedestrian detectors, we verify the adaptability and effectiveness of our method showing noticeable performance gains and achieving state-of-the-art detection performance on two public pedestrian detection benchmarks (\textit{i.e.,} \textit{CrowdHuman} and \textit{WiderPedestrian}).
\end{abstract}


%
\IEEEpeerreviewmaketitle

 

\section{Introduction} \label{intro}
%
%
%
%


\makeatletter
\def\ps@IEEEtitlepagestyle{
  \def\@oddfoot{\mycopyrightnotice}
  \def\@evenfoot{}
}
\def\mycopyrightnotice{
  {\footnotesize
  \begin{minipage}{\textwidth}
  \centering
    Copyright~\copyright~2024 IEEE. Personal use of this material is permitted. However, permission to use this \\ material for any other purposes must be obtained from the IEEE by sending an email to pubs-permissions@ieee.org.
  \end{minipage}
  }
}

\IEEEPARstart{T}{hese} days, large language models (LLMs) have emerged and accelerated the evolution of deep learning, showcasing their exceptional capabilities in contextual understanding, interpretability, generalizability, and so on \cite{context1, gpt-3, ctrl, bert, context2}. Even though LLMs are initially designed to handle and revolutionize natural language processing (NLP) mainly, it would be effective to exploit their strengths for computer vision models in many aspects.

There have been several attempts to take advantage of LLMs in computer vision tasks \cite{csvt_lang1, csvt_lang2, vild, labo, i2mvformer, lavila, visionllm}. For example, Yang \textit{et al.} \cite{labo} leveraged GPT-3 \cite{gpt-3} to design an interpretable image classifier by generating conceptual descriptions of given images and comparing them with image features. Naeem \textit{et al.} \cite{i2mvformer} used PaLM \cite{palm} to provide text supervision for zero-shot image classification. Shao \textit{et al.} \cite{shao-vqa} tried to prompt GPT-3 heuristically and obtain the final answer. Zhao \textit{et al.} \cite{lavila} exploited T5 \cite{t5} and generated pseudo descriptions for videos to augment video and narration pair data.

\begin{figure}[t]
    \begin{center}
    \centerline{\includegraphics[width=0.999\linewidth]{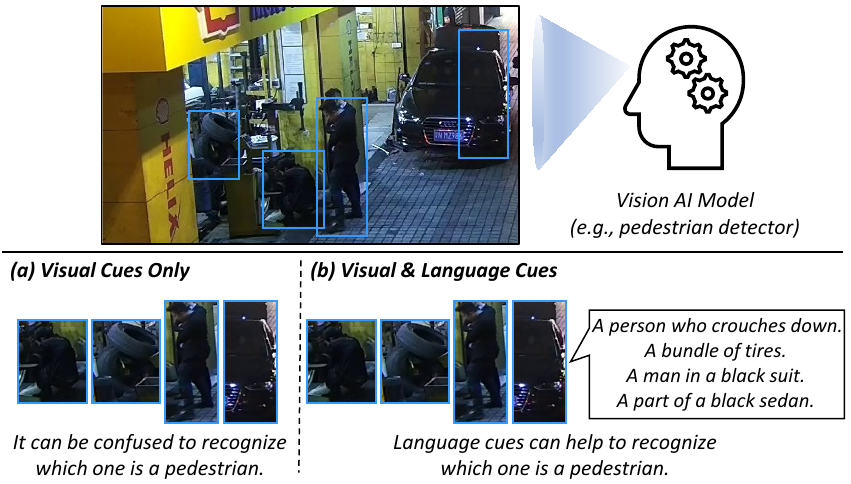}}
    \end{center}
    \caption{This illustrates that a vision AI model (\textit{i.e.,} pedestrian detector) recognizes pedestrians in the street scene. (a) When visual cues are provided only, the pedestrian detector could be confused about recognizing instances because of their similar appearances (\textit{e.g.,} color). (b) However, when language cues are provided (\textit{e.g.,} ``\textit{A person who crouches down.}''), they can help the detector to perceive instances properly.}
\label{fig1}
\end{figure}

It is also known that LLMs have demonstrated their ability in understanding appearance information \cite{gpt-3, labo}. Therefore, it is very intuitive to consider that auxiliary language knowledge of appearances obtained from LLMs can help a vision model to perceive visual scenes and instances. As shown in Fig. \ref{fig1}, several instances, such as pedestrians, a bunch of tires, and a vehicle, appear in the given visual scene. When a vision model (here, a pedestrian detector) tries to recognize pedestrians, it can be confusing to identify them with visual cues only (Fig. \ref{fig1}(a)). However, appearance descriptions can provide auxiliary information (\textit{e.g.,} pose), so that it would help the vision model to perceive instances properly.

In this paper, we introduce a novel approach to take advantage of LLMs understanding contextual appearance variations and to utilize the appearance knowledge they understand into a vision model, that is pedestrian detection. Pedestrian detection is one of the critical tasks, which is directly involved in our safety (\textit{e.g.,} intelligent driving and surveillance systems) \cite{csvt_ped1, csvt_ped2, csvt_ped3, csvt_ped4, csvt_ped5, csvt_ped6}. However, it is considered very challenging, because there are a variety of appearance variations, such as pose and direction, in diverse scenes. Therefore, we attempt to handle such problems by exploiting the strength of LLMs in understanding appearance contexts. To this end, we compose a description corpus consisting of abundant narratives describing different appearances of pedestrians and other instances. Then we feed them into an LLM and obtain the description embeddings, named \textit{appearance knowledge sets}. The appearance knowledge sets are comprised of plentiful appearance representations of instances. Among the huge knowledge sets, we sample representative appearance knowledge, called \textit{appearance knowledge centroids}. We repurpose them via task-prompting and guide them to be relevant to a downstream pedestrian detection task. By doing so, we acquire \textit{appearance elements} which are representative appearance knowledge and task-related at the same time. Moreover, when appearance elements are obtained, they are off-the-shelf, so that they can be adaptable to various detection frameworks. Therefore, we can incorporate the appearance elements with instances' visual representations in a pedestrian detector.

Different from the previous methods adopting LLMs to generate texts and compare them with image features mainly or to augment data, in this work, we present a novel approach to formulate language-derived appearance elements and to integrate them with visual cues. For the integration, any language input or text generation processes are not required during the inference phase. In other words, it does not require LLMs for the model inference. Thus, the obtained language-derived appearance elements can be applicable to various detection frameworks and diverse visual scene data. Through comprehensive experiments with various pedestrian detection frameworks, we validate the adaptability and effectiveness of the proposed method achieving state-of-the-art performance with remarkable performance improvement.

Our contributions can be summarized as follows:
\begin{itemize}
    \item We present a novel method of taking advantage of LLMs to provide plentiful appearance information to a pedestrian detector, so that the pedestrian detector benefits in perceiving visual scenes and instances more effectively. As far as we know, this is the first work to leverage language-derived appearance knowledge and to incorporate it with visual cues in pedestrian detection.
    \item We establish a description corpus that is composed of numerous descriptions illustrating diverse views of instances. We also formulate off-the-shelf appearance elements that contain representative language-derived appearance knowledge and are guided to be task-related via task-prompting. Then, we integrate them with instances' visual representations even without any additional language inputs during the inference time.
    \item Through extensive experiments with various detection frameworks, we corroborate its adaptability and effectiveness on public pedestrian detection benchmarks, \textit{CrowdHuman} \cite{crowdhuman} and \textit{WiderPedestrian} \cite{wider}, showing state-of-the-art performance along with a large performance gain.
\end{itemize}

\begin{figure*}[t]
    \begin{center}
    \centerline{\includegraphics[width=0.85\linewidth]{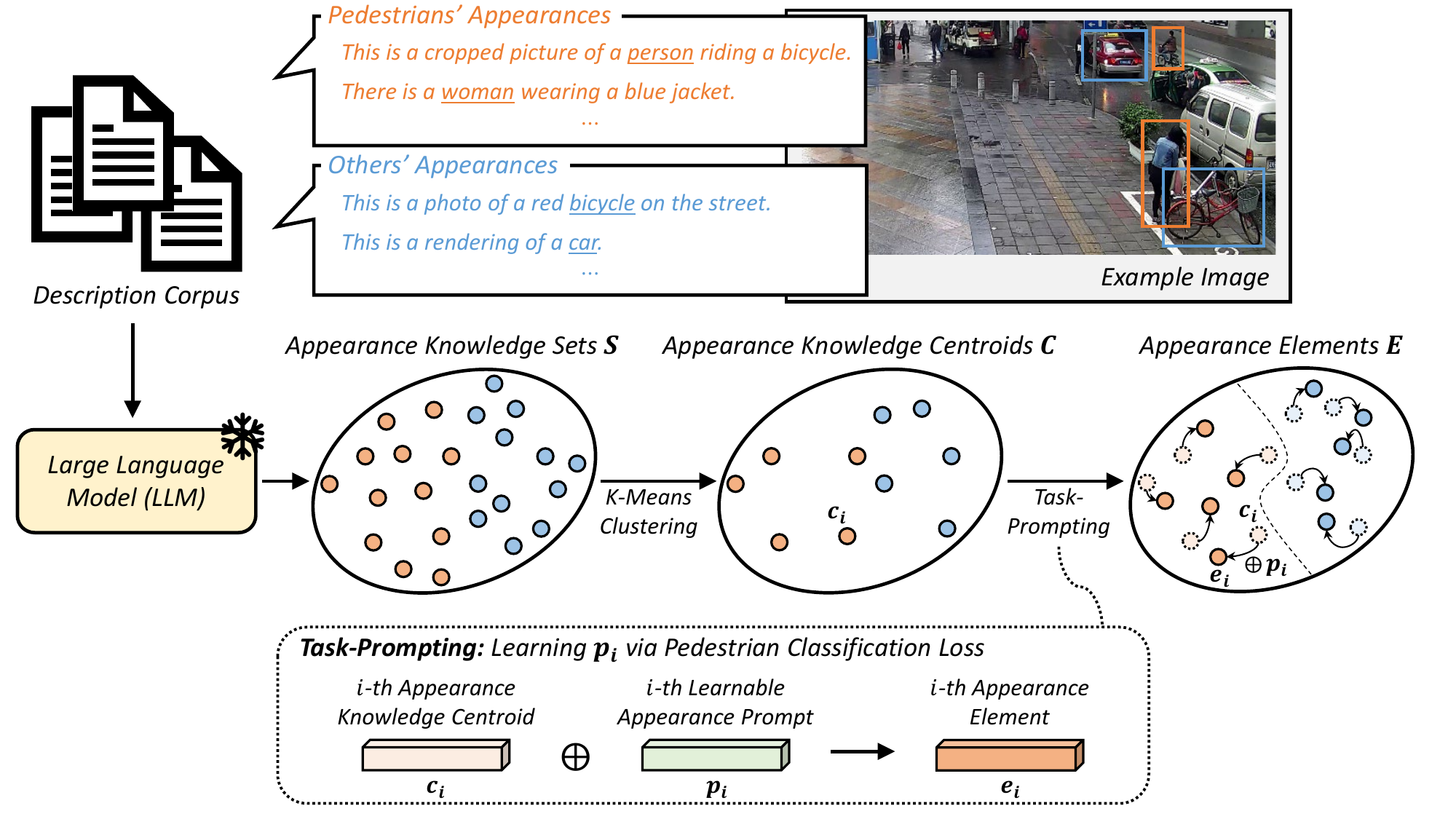}}
    \end{center}
    \caption{The overview of formulating appearance elements through an LLM. After composing a description corpus that includes diverse appearance variations of instances, an LLM takes them to obtain the appearance knowledge sets $\boldsymbol{S}$, which contain a number of appearance representations. Among the numerous appearance features in the knowledge set $\boldsymbol{S}$, we sample $K$ numbers of representative appearance knowledge, named appearance knowledge centroids $\boldsymbol{C}$, by using K-means clustering. Along with $K$ centroids, we place $K$ learnable appearance prompts $\boldsymbol{P}$, and we conduct an elementwise summation with $\boldsymbol{C}$ and $\boldsymbol{P}$. Therefore, we obtain appearance elements $\boldsymbol{E}$. Furthermore, we repurpose $\boldsymbol{E}$ with pedestrian classification loss which is directly related to the pedestrian detection downstream task, that is task-prompting. During task-prompting, $\boldsymbol{P}$ is the only learnable parameters updated only. Therefore, we can formulate $\boldsymbol{E}$ which contains appearance knowledge from an LLM and becomes more task-relevant.}
\label{fig2}
\end{figure*}

\section{Related Work}
\subsection{Large Language Models (LLMs)}
Large language models (LLMs) \cite{llm1, llm2, llm3, llm4, llm5}, such as BERT \cite{bert}, GPT \cite{gpt}, CTRL \cite{ctrl}, and T5 \cite{t5}, are usually the models that are trained with a huge web-scale database, and they have proved their strong capabilities and potentials. BERT \cite{bert} is one of the fundamental LLMs, and it has accelerated the evolution of natural language processing (NLP) by obtaining noticeable performances on various downstream tasks. CTRL \cite{ctrl} was presented to focus on language generation conditioned on style, content, and task-specific behavior. Raffel \textit{et al.} \cite{t5} explored transfer learning techniques for NLP and presented a versatile framework named T5. It regards various text-based language tasks as text-to-text problems. Although such LLMs are initially designed for NLP tasks, they have demonstrated their exceptional capabilities in many aspects (\textit{e.g.,} contextual understanding). In this paper, we concentrate on their intelligibility in understanding appearance descriptions, and we incorporate this knowledge with visual cues of a pedestrian detector.

\subsection{Vision Models with LLMs}
As LLMs have shown noticeable capabilities in many aspects, several methods have attempted to exploit their strengths in vision models \cite{vild, clip, vilbert, visionllm}, and the various vision models have benefited from the advantages of language modalities \cite{csvt_vis_lang1, csvt_vis_lang2, csvt_vis_lang3, csvt_vis_lang4}. Yang \textit{et al.} \cite{labo} proposed LLM-guided concept bottlenecks for an interpretable image classification framework, named LaBo. LaBo exploited GPT-3 \cite{gpt-3} to obtain candidate concepts and calculate the similarities with a testing image. Naeem \textit{et al.} \cite{i2mvformer} introduced a zero-shot image classification framework by generating documents with PaLM \cite{palm}, and then, they compared text features with image features to find out which category a testing image belongs to. Shao \textit{et al.} \cite{shao-vqa} presented Prophet which is a visual question answering (VQA) framework adopting GPT-3 to predict better answers. After a vanilla VQA model generates answer heuristics, they are combined with the given question and caption prompts to be fed into GPT-3. Zhao \textit{et al.} \cite{lavila} tried to enhance video representations by exploiting T5 \cite{t5} for data augmentation. T5 is used to narrate videos, and the obtained narration is used to augment video and text pair data. Most of the methods that exploit an LLM usually concentrate on 1) generating documents with an LLM, 2) comparing text features with image features, or 3) augmenting data. However, different from the previous works, we focus on integrating language appearance cues with visual cues in a vision model (\textit{i.e.,} pedestrian detection), so that it encourages the vision model to perceive instances well. Furthermore, the proposed method does not require any language input or text generation from an LLM during the inference time.

\subsection{Pedestrian Detection}
Pedestrian detection is considered one of the crucial computer vision tasks, because it is directly related to real-world safety applications, such as intelligent driving and surveillance systems \cite{ped3, ped1, ped4, ped5, deng_aaai, csvt_ped7, csvt_ped8, csvt_ped9, csvt_ped10}. Li \textit{et al.} \cite{li2017adaptive} proposed to build a surveillance scene specific detector by selecting convolutional kernels adaptively for accurate pedestrian localization. Lin \textit{et al.} \cite{csvt_ped5} proposed a multi-grained deep feature learning (MDFL) that jointly trains a multi-scale network and a human parsing network. Wang \textit{et al.} \cite{csvt_ped6} explored the relations between the scales and prediction scores of pedestrians and figured out the impact on prediction scores by pedestrians' scales. Jiao \textit{et al.} \cite{csvt_ped1} presented a pose embedding network (PEN) consisting of dual paths identifying pedestrians and recognizing their poses. CFRLA-Net \cite{csvt_ped4} was introduced to integrate convolution and multi-head self-attention paths to capture both local and global spatial contexts. Zhang \textit{et al.} \cite{ped2} designed attribute-NMS to remove false positives in crowded scenes and devised pedestrian oriented attribute maps to attend pedestrian features. Zheng \textit{et al.} \cite{e2edet} introduced an iterative pedestrian detection framework named E2EDET. It iterates accept and reject process, selecting only object queries with high scores. Moreover, Zhang \textit{et al.} \cite{ddq} observed that a traditional NMS is not appropriate for a sparse query algorithm, and they introduced dense distinct queries (DDQ) to acquire more accurate queries. In this work, our method is designed to provide visual cues with contextual appearance variation knowledge, not restricted to a particular scene, by 1) utilizing numerous narratives describing diverse views of instances and 2) formulating language-derived appearance elements with the help of the contextual understanding capability of an LLM.

Furthermore, Tang \textit{et al.} \cite{tang2023otp} designed OTP-NMS which predicts the optimal threshold of NMS depending on the visibility ratio and classification score for robust pedestrian detection. Tu \textit{et al.} \cite{tu2018semantic} used an additional flow modality and divided the upper and lower parts of a person to capture a person’s accurate action. Cao \textit{et al.} \cite{cao2021handcrafted} conducted a comprehensive investigation on recent pedestrian detection works and categorized them depending on their research directions. Liu \textit{et al.} \cite{vlpd} proposed VLPD which utilizes CLIP to obtain pseudo prototypes of other categories for pixel level contrastive learning. Different from the previous methods \cite{tang2023otp, tu2018semantic}, which mainly concentrated on improving the post-processing algorithms or encoding visual parts separately, the proposed method is designed to formulate language-derived appearance elements and incorporate them with visual cues of a pedestrian. Based on \cite{cao2021handcrafted}, unlike the recent research direction in pedestrian detection (\textit{e.g.,} part-based or post-processing enhancement), our method presents an innovative approach utilizing language and complementing visual cues. Furthermore, while VLPD \cite{vlpd} was proposed to pull away pedestrian representations from others, our method is designed to provide plentiful fine-grained appearance knowledge and formulate off-the-shelf knowledge elements by utilizing an LLM.

\section{Proposed Method}
In this section, we introduce the details of the proposed method: 1) how to compose appearance description corpus, 2) how to formulate language-derived appearance elements, and 3) how to integrate the appearance elements with visual cues in a pedestrian detector. The overview of how to prepare the appearance elements is illustrated in Fig. \ref{fig2}. After establishing description corpus composed of diverse appearance illustrations, we feed them into an LLM to acquire appearance knowledge sets that contains appearance description embeddings. Then we formulate appearance elements via K-means clustering and task-prompting. The details of the proposed method are described in the following subsections.

\subsection{Appearance Description Corpus}
The purpose of appearance description corpus is to express diverse appearances of numerous instances. In pedestrian detection, there are two categories to be recognized, \textit{pedestrian} and \textit{background}, where \textit{background} includes any other instances except pedestrian. Therefore, we compose description corpus that contains millions of appearance views for pedestrian and other instances. Then we extract huge appearance knowledge sets by feeding them into an LLM. The details and basic ways generating templates for both pedestrian and background are described below.

\begin{itemize}
    \item Hand-crafted templates: We adopt widely used hand-crafted template formats \cite{clip}, and we curate them following \cite{vild}. The examples of template formats are `\textit{There is \{article\} \{class\} in the scene.}', `\textit{A photo of \{article\} \{class\}.}', \etc. Here, \textit{\{article\}} is one of \textit{\{a/an/the\}}, and \{class\} denotes the class name, such as pedestrian.
    \item Word variations: We also consider word variations by substituting the class name with its synonyms based on WordNet \cite{wordnet, wordnet2} to acquire varying descriptions with similar meaning. For instance, human can express pedestrian in many other words, such as person, man/woman, boy/girl, and so on.
\end{itemize}

\begin{table}[!t]
	\renewcommand{\arraystretch}{1.3}
	\renewcommand{\tabcolsep}{3.5mm}
\centering
\caption{The examples of attribute adjectives used to obtain pedestrian appearance descriptions.}
\resizebox{0.9\linewidth}{!}{
\begin{tabular}{cc}
\Xhline{3\arrayrulewidth}
\textbf{Types} & \textbf{Attribute Examples} \\ \hline
Age & young, old, little, elderly, ... \\
Body & tall, short, big, small, ... \\
Expression & smiling, crying, displeased, ... \\
Clothes (hair) & t-shirt, dress, jeans, hat, hair, ... \\
Color & white, black, red, blue, ... \\
Pose & standing, walking, sitting, crouching, ... \\
Direction & in front, in profile, from behind, ... \\
Action & riding a bicycle, playing a baseball, ... \\ \Xhline{3\arrayrulewidth}
\end{tabular}}
	\label{tab1}
\end{table}

\begin{table}[!t]
	\renewcommand{\arraystretch}{1.3}
	\renewcommand{\tabcolsep}{3.5mm}
\centering
\caption{The examples of appearance descriptions for pedestrian and background instances.}
\resizebox{0.99\linewidth}{!}{
\begin{tabular}{c}
\Xhline{3\arrayrulewidth}
\textbf{Appearance Description Examples (Pedestrian)} \\ \hline
A photo of a tall lady wearing a red backpack. \\
A cropped picture of a fat man with eyeglasses. \\
There is a slim person in the scene. \\
A rendering of a slim woman wearing a white hat. \\
A blurry rendering of a young guy wearing blue pants. \\
A dark rendering of a child with yellow hair. \\
A blurry rendering of a short stroller wearing black clothes. \\
A close-up painting of a big boy in a gray. \\
A bright painting of a young player playing a basketball. \\
itap of a thin commuter wearing a green t-shirt. \\
A rendering of a tall man wearing a black jacket. \\
A photo of the hiker from behind. \\
A cropped painting of a main in front. \\
A picture of a thin woman with gray hair in the scene. \\
A rendering of a slim lady wearing black pants. \\
A blurry picture of a short person exercising. \\
A close-up rendering of a big guy riding a bike. \\
A low resolution photo of a girl playing a baseball. \\
A cropped photo of a big player wearing black sunglasses. \\
A blurry photo of a short boy playing a guitar. \\
... \\ 
\\ \hline \hline
\textbf{Appearance Description Examples (Others)} \\ \hline
A dark rendering of a truck. \\
A painting of the small motorcycle. \\
A low resolution rendering of the umbrella. \\
A cropped picture of the car. \\
A black and white photo of the cat. \\
A low resolution painting of a small car. \\
A photo of the hard to see vehicle. \\
A black and white photo of the street lamp. \\
A picture of the dirty truck. \\
A rendering of the large tree. \\
Cute wedding background with roses, lace and place for text Illustration. \\
The concrete mixer truck. \\
This image shows a truck and car double parking in NYC. \\
basketball ball. \\
A fire hydrants dug out from the snow. \\
Solar led street light. \\
outdoor tennis court. \\
River deep in mountain and forest - Stock Photo. \\
Image of a black motorcycle, illustrating car sweepstakes at About.com. \\
A tiny dog in a basket that was connected to a motor bike. \\
... \\ 
\\ \Xhline{3\arrayrulewidth}
\end{tabular}}
	\label{tab2}
\end{table}

\subsubsection{Pedestrian Templates} We first omit the ambiguous adjectives used in the basic hand-crafted templates \cite{vild} to suit out purpose. For example, we remove `nice' and `cool' which are considered inappropriate to describe pedestrians' appearance. So that, several basic templates, such as `\textit{A photo of a nice \{class\}.}', `\textit{A photo of a cool \{class\}.}', are merged into one same template, that is `\textit{A photo of \{article\} \{class\}.}'. We also discard the templates that are not proper for real-world pedestrian detection, for example, `\textit{The plastic \{class\}.}', `\textit{A toy \{class\}.}'. With residual basic hand-crafted templates, we give more specified attribute adjectives to generate plentiful appearances of pedestrian, as shown in Table \ref{tab1}. The types of `Age', `Body', and `Expression' are located in the front of \{\textit{class}\}. The others are positioned in the following of \{\textit{class}\}. Each of adjectives is independently decided to be used or not with a $0.5$ probability. `Color' attribute is usually utilized in the manner of `in \{\textit{article}\} \{\textit{Color}\} \{\textit{Clothes}\}', and also `Clothes' can be used with prefixes like `\textit{in}, \textit{wearing}, and \textit{with}, and so on. Finally, the final format of description for pedestrian becomes `\textit{\{template\} \{article\} \{age/body/expression\} \{class\} \{clothes/color/pose/direction/action\}.}' where \textit{\{template\}} denotes one of the hand-crafted templates. Several examples are shown in Table \ref{tab2}. 

\subsubsection{Background Templates} For background instances, we collect the classes irrelevant from pedestrian among the categories of MS-COCO \cite{coco}, such as \textit{car}, \textit{dog}, \textit{potted plant}, \etc. To describe these instances, we use the `Color' attribute only in front of \textit{\{class\}}. In other words, the attribute adjectives, such as pose and action, are not used, because the background categories include infinite and diverse kinds of objects which are not matched well with such attributes. For example, expression attributes (\eg \ `smiling' and `crying') are unsuitable for describing `tree', `vehicle', and so on. Therefore, we just use the basic hand-crafted \cite{vild} without any curation used for pedestrian templates. We also supplement the background descriptions by using LAION dataset \cite{laion}. We remove text data which are related to pedestrian and use the residuals for background descriptions illustrating extensive contexts. As a result, we establish millions of descriptions for illustrating appearance variations of each pedestrian and background. Several examples are also shown in Table \ref{tab2}.

\subsection{Language-Derived Appearance Elements}
In this subsection, we introduce the way to extract language-derived appearance elements with the constructed description corpus which contains diverse appearance illustrations of many instances. Fig. \ref{fig2} shows the overview of how to generate appearance elements. First, an LLM takes description corpus and extracts description sentence embedding sets, named \textit{appearance knowledge sets} $\boldsymbol{S} = \{\boldsymbol{s}_{j}\}^M_{j=1}$. It consists of the plentiful appearance representations of instances. Here, $\boldsymbol{s}_{j} \in \mathbb{R}^{d}$ is the $j$-th knowledge embedding, and $M$ is the number of extracted knowledge embeddings as same as the number of description corpus. Second, we perform K-means clustering with $\boldsymbol{S}$ to sample $K$ representative appearance knowledge, called \textit{appearance knowledge centroids} $\boldsymbol{C}=\{\boldsymbol{c}_{j}\}^{K}_{j=1}$ where $\boldsymbol{c}_{j} \in \mathbb{R}^{d}$. Then we find out which centroid each appearance knowledge embedding $\boldsymbol{s}_{j}$ belongs to, as follows:
\begin{equation}
    i = \underset{k \in [1, K]}{argmax} \, \boldsymbol{s}_{j} \cdot \boldsymbol{c}_{k},
    \label{eq1}
\end{equation}

\noindent
where $i$ is the index that the $j$-th appearance embedding $\boldsymbol{s}_{j}$ belongs to among $\boldsymbol{C}$, and $\boldsymbol{s}_{j}$ can be substituted by the corresponding centroid $\boldsymbol{c}_{i}$. 

Next, we repurpose $\boldsymbol{C}$ via task-prompting to make them related to a downstream pedestrian detection task. As several methods observed that the representation from large pretrained models, such as $\boldsymbol{C}$ can be usually sub-optimal because of its semantic gap with downstream tasks \cite{adaprompt, vt-clip}, it is motivated to place learnable parameters to mitigate such a gap and adapt the obtained representation to be more task-relevant. To this end, we place learnable appearance prompts $\boldsymbol{P}=\{\boldsymbol{p}_{j}\}^K_{j=1}$ where each of them is paired with the centroid having the same index. While adding $\boldsymbol{P}$ with the corresponding $\boldsymbol{C}$ in elementwise and resulting \textit{appearance elements}, we guide them with pedestrian classification loss which is directly associated with pedestrian detection. Specifically, we obtain \textit{appearance elements} $\boldsymbol{E}=\{\boldsymbol{e}_{j}\}^K_{j=1}$ as follows:
\begin{equation}
    \boldsymbol{e}_{i} = \boldsymbol{c}_{i} \oplus \boldsymbol{p}_{i}.
    \label{eq2}
\end{equation}

\noindent
As mentioned above, we train the prompt $\boldsymbol{p}_{i}$ with a pedestrian classification loss (\ie \ binary cross-entropy loss). Depending on the instance type of the given $\boldsymbol{s}_{j}$ in Equation (\ref{eq1}), it guides $\boldsymbol{e}_{i}$ to either pedestrian or background. By doing so, we can acquire $\boldsymbol{E}$ consisting of: 1) the representatives of language-derived appearance knowledge, and 2) task-relevant features fit to distinguish pedestrian and others.

The processes of training the learnable appearance prompts $\boldsymbol{P}$ and obtaining appearance elements $\boldsymbol{E}$ are performed before the training of a pedestrian detection framework. After the processes, the appearance elements $\boldsymbol{E}$ are integrated into a pedestrian detection framework, and the details are described in the following subsection.

\begin{figure}[t]
    \begin{center}
    \centerline{\includegraphics[width=0.85\linewidth]{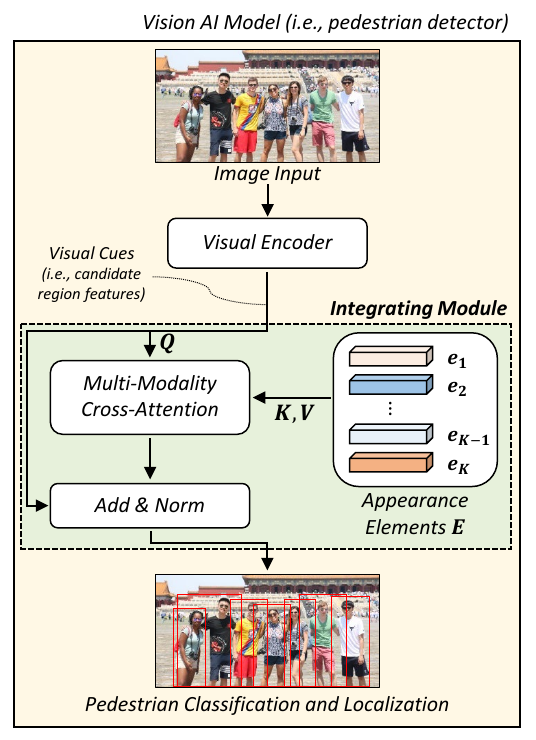}}
    \end{center}
    \caption{The way to incorporate language-derived appearance elements $\boldsymbol{E}$ with visual features in a pedestrian detector. The integrating module, consisting of multi-modality cross-attention, addition (\textit{Add}), and normalization(\textit{Norm}), is embedded into a pedestrian detector. The cross-attention takes visual query features ($\boldsymbol{Q}$) to refer to $\boldsymbol{E}$ as key ($\boldsymbol{K}$) and value features ($\boldsymbol{V}$).}
\label{fig3}
\end{figure}

\subsection{How to Integrate Appearance Elements with Visual Cues}
After establishing language-derived appearance elements $\boldsymbol{E}$, it is required to integrate $\boldsymbol{E}$ with visual cues in a pedestrian detector. Fig. \ref{fig3} illustrates the overview, showing the integrating module embedded within a pedestrian detector. When an input image is given, a visual encoder extracts visual cue features. Please note that, in pedestrian detection, visual cue features are considered as candidate region features that can be either pedestrian or others (\eg \ region of interests (RoIs) and object query features). As shown in the green box of the figure, the integrating module is composed of multi-modality cross-attention module, addition, and normalization (\textit{Add \& Norm}) \cite{vilbert, lxmert, att}. We utilize the cross-attention to take visual cue features as query ($\boldsymbol{Q}$) and appearance elements $\boldsymbol{E}$ as key ($\boldsymbol{K}$) and value ($\boldsymbol{V}$), so that we aggregate visual region features and language-derived appearance knowledge features. After \textit{Add \& Norm}, we obtain the incorporated features which are used to perform the final pedestrian detection (\ie \ classification and localization). On the one hand, a visual query feature $\boldsymbol{q} \in \boldsymbol{Q}$ is a region feature that can be either pedestrian or background. Therefore, during the training phase, we guide a visual query feature $\boldsymbol{q}$ to refer to the proper elements which are related to the corresponding category for the cross-attention. Specifically, when a pedestrian candidate query feature comes in, it requires to read the elements that contain pedestrian's appearance knowledge not background. Hence, we make attention scores between a pedestrian candidate query feature and background-related elements $\boldsymbol{E}_{b} \subset \boldsymbol{E}$ be small. In the opposite case, we guide the attention scores between a background candidate query feature and pedestrian-related elements $\boldsymbol{E}_{p} \subset \boldsymbol{E}$ to be low, and the reference loss $\mathcal{L}_{ref}$ is designed as follows:
\begin{equation}
    \begin{aligned}
        \mathcal{L}_{ref} = {\mathbbm{1}_{ped} \over |\boldsymbol{E}_{b}|} \sum_{i \in \boldsymbol{E}_{b}} softmax(\boldsymbol{q} \cdot \boldsymbol{E})_i\\+ {1 - \mathbbm{1}_{ped} \over |\boldsymbol{E}_{p}|} \sum_{j \in \boldsymbol{E}_{p}} softmax(\boldsymbol{q} \cdot \boldsymbol{E})_j.
    \end{aligned}
    \label{eq3}
\end{equation}

\noindent
During the training time, we use an indicator $\mathbbm{1}_{ped}$ which is assigned by $1$ for pedestrian and $0$ for background to identify whether a query feature is a pedestrian or not. In Equation (\ref{eq3}), the first term makes the attention scores between the given pedestrian query feature and $\boldsymbol{E}_{b}$ to be small, and the second term is for the opposite case. Although it seems that $\mathcal{L}_{ref}$ considers the relationship between negative pairs only, we can expect that residual scores, that is the attention scores between positive pairs, become higher at the same time due to the softmax function. By designing $\mathcal{L}_{ref}$, we help visual query features to refer to the appropriate knowledge elements of the corresponding instance (pedestrian or others). Finally, we incorporate $\mathcal{L}_{ref}$ with the existing pedestrian detection losses (\textit{i.e.,} focal loss, L1 loss, and GIoU loss).

\section{Experiments}
\subsection{Pedestrian Detection Benchmarks}
\subsubsection{CrowdHuman \cite{crowdhuman}} It is one of the widely-known pedestrian detection benchmarks, and the images are collected from huge web-crawling data. Therefore, it contains the images collected in various environments, such as indoor places, outdoor squares, urban and country places, \etc. It is composed of 15,000, 4,370, and 5,000 images for each training, validation, and test sets, respectively, and it contains around 340,000 pedestrian instances in the training set. For the fair evaluation and comparison, we use a full-body annotation and validation set following the previous works \cite{e2edet, deng_aaai}. \textit{Please note that, different from some methods, we do not use any extra data to train pedestrian detection frameworks except \textit{CrowdHuman}}.
\subsubsection{WiderPedestrian \cite{wider}} This benchmark is another large pedestrian detection benchmark, and we use it to validate the proposed method in the safety environment. It consists of 11,500 training, 5,000 validation, and 3,500 testing images which are collected from driving and surveillance scenes for the safety of pedestrians. For both benchmarks, we adopt average precision with IoU threshold 0.5 (we denote it `AP' for simplicity) for the evaluation, as the previous works have adopted it as the primary evaluation metric \cite{e2edet}. The higher AP indicates the better detection performance. As the testing annotations are inaccessible, we use the validation set for the evaluation.

\subsection{Implementation Details}
\subsubsection{Method Details} When we compose a description corpus which contains diverse appearances of instances, we collect 1.2M descriptions for pedestrian and 10M for background. For obtaining the corpus with WordNet \cite{wordnet2}, we use python NLTK library \cite{nltk}. Then we randomly sample 50,000 descriptions from each of them to feed them into an LLM and extract appearance knowledge embeddings. Note that the random sampling process is performed without any replacement, so that once a description is selected, it cannot be selected again. The total number of appearance knowledge sets, $M$, is 100,000. We use Sentence-T5 \cite{sentence-t5} to extract description embeddings (not token outputs), which explored T5 \cite{t5} to obtain sentence embeddings for processing language tasks effectively, and the dimension of appearance knowledge, $d$, is 768 dimensions. Among 100,000 appearance knowledge sets, we extract $K=200$ appearance knowledge centroids as a default, so that the number of language-derived appearance elements is also $200$. To repurpose the appearance elements during task-prompting, we train the learnable appearance prompt. The learnable prompt is element-wise added with the knowledge centroid, and then the appearance element is obtained. We guide the learnable prompt by using binary cross entropy (BCE) objective with learning rate $0.1$ to predict whether the appearance element is a pedestrian or not. Regarding the architecture, we adopt classifier head architecture composed of two linear layers (one for projection and one for classification), and we feed the element into this classifier head. This classifier head architecture for training the prompt is based on the classification layers which are widely used in detection frameworks (\textit{e.g.,} Sparse R-CNN \cite{sparse}).

\subsubsection{Model and Training Details} Regarding pedestrian detection frameworks, we adopt Sparse R-CNN \cite{sparse}, D-DETR \cite{d-detr}, DDQ R-CNN, and DDQ DETR \cite{ddq} which are the representatives of widely-used and latest detection frameworks. We use PyTorch library to implement the detection frameworks. We follow the training protocols described in each method as described below:
\begin{itemize}
    \item \textbf{Sparse R-CNN} \cite{sparse, e2edet} - \textit{Epoch:} $50$, \, \textit{Optimizer:} AdamW, \, \textit{Learning rate:} $5e-5$, \, \textit{Weight decay:} $1e-4$, \, \textit{Warm-up factor:} $1e-2$, \, \textit{Warm-up iteration:} $1,000$
    
    \item \textbf{D-DETR} \cite{d-detr, e2edet} - \textit{Epoch:} $50$, \, \textit{Optimizer:} AdamW, \, \textit{Learning rate:} $2e-4$, \, \textit{Weight decay:} $1e-4$, \, \textit{Warm-up factor:} $1e-2$, \, \textit{Warm-up iteration:} $1,000$
    
    \item \textbf{DDQ R-CNN} \cite{ddq} - \textit{Epoch:} $36$, \, \textit{Optimizer:} AdamW, \, \textit{Learning rate:} $1e-4$, \, \textit{Weight decay:} $5e-2$, \, \textit{Warm-up factor:} $1e-4$, \, \textit{Warm-up iteration:} $2,000$
    
    \item \textbf{DDQ DETR} \cite{ddq} - \textit{Epoch:} $36$, \, \textit{Optimizer:} AdamW, \, \textit{Learning rate:} $2e-4$, \, \textit{Weight decay:} $5e-2$, \, \textit{Warm-up factor:} $1e-4$, \, \textit{Warm-up iteration:} $2,000$
\end{itemize}
\noindent
For the cross-attention in the integrating module, we build it as 8 multi-head attention architectures as a default, and we use a single-head attention architecture for DDQ DETR. We use 8 NVIDIA RTX A6000 GPUs for the training. Our code is available at: \MYhref[magenta]{https://github.com/kimhj709/LDAE}{\texttt{https://github.com/kimhj709/LDAE}}.

\subsection{Comparisons of Pedestrian Detection Results}
On \textit{CrowdHuman} and \textit{WiderPedestrian} benchmarks, we compare the proposed method with state-of-the-art pedestrian detection frameworks \cite{sparse, d-detr, e2edet, pmip, deng_aaai, ddq}. Table \ref{tab3} shows performance comparison results on \textit{CrowdHuman}. We apply the proposed method into four detection frameworks to show its effectiveness and adaptability, which are Sparse R-CNN, D-DETR, DDQ R-CNN, and DDQ DETR. Compared to the baselines for each of them, our method obtains large performance gains, 1.9AP, 2.7AP, 1.0AP, and 0.6AP, respectively, and it achieves state-of-the-art detection performance outperforming the existing methods. Note that, while DDQ DETR \cite{ddq} used ResNet-50 backbones \cite{resnet} in the original paper achieving 93.8AP, we reimplement it with Swin-L backbones \cite{swin} for both baseline (94.8AP) and our method (95.4AP). Moreover, Table \ref{tab4} shows the comparison with the existing methods on \textit{WiderPedestrian}. Since this benchmark is composed of safety environments which are driving and surveillance, it can show the effectiveness of our method on such environments more properly. We adopt three detection frameworks, and they obtain noticeable performance improvements achieving state-of-the-art detection performances. Through these experiments, we can corroborate that: 1) the proposed method can be adopted in various pedestrian detection frameworks, and 2) the obtained appearance knowledge elements $\boldsymbol{E}$ can be exploited in diverse scenes. Note that \textit{CrowdHuman} is composed of various scene data from web-crawling, and \textit{WiderPedestrian} consists of driving and surveillance scene data mainly.

\begin{table}[t]	
	\centering
    \caption{The comparison with the existing methods on \textit{CrowdHuman}. `\textbf{\# Queries}' denotes the number of object queries in a pedestrian detector. We adopt average precision (AP) as evaluation metric. With four pedestrian detection frameworks, the proposed method obtains large performance gains, while achieving state-of-the-art detection performance on \textit{CrowdHuman}.}
	\begin{center}
		\renewcommand{\tabcolsep}{3.5mm}
		\resizebox{0.99\linewidth}{!}
		{
			\begin{tabular}{ccc}
				\Xhline{3\arrayrulewidth}
                    \rule{0pt}{10pt}
				    \bf Method & \bf \# Queries & \bf AP \\ \hline
                    \rule{0pt}{10pt}
                    Sparse R-CNN (CVPR'21) & 500 & 90.7 \\
                    D-DETR (ArXiv'20) & 1000 & 91.5 \\
                    E2EDET (Sparse R-CNN) (CVPR'22) & 500 & 92.0 \\
                    E2EDET (D-DETR) (CVPR'22) & 1000 & 92.1 \\                    
                    DRFG + PMIP (PR'22) & - & 92.2 \\
                    Deng \etal \ (AAAI'23) & 500 & 92.3 \\
                    DDQ R-CNN (CVPR'23) & 300 & 93.5 \\
                    DDQ DETR (CVPR'23) & 900 & 94.8 \\ \hline
                    \rule{0pt}{10pt}
                    \bf Ours (Sparse R-CNN) & \bf 500 & \bf 92.6 \\
                    \bf Ours (D-DETR) & \bf 1000 & \bf 94.2 \\
                    \bf Ours (DDQ R-CNN) & \bf 300 & \bf 94.5 \\
                    \bf Ours (DDQ DETR) & \bf 900 & \bf 95.4 \\ \Xhline{3\arrayrulewidth}
			\end{tabular}
		}
	\end{center}
	\label{tab3} 
\end{table}

\begin{table}[t]
	\centering
    \caption{The comparison with state-of-the-art methods on \textit{WiderPedestrian}. `\textbf{\# Queries}' denotes the number of object queries in the detection framework. We adopt average precision (AP) as evaluation metric. With three pedestrian detection frameworks, the proposed method obtains noticeable performance improvements, achieving state-of-the-art detection performance on \textit{WiderPedestrian}.}
	\begin{center}
		\renewcommand{\tabcolsep}{3.5mm}
		\resizebox{0.99\linewidth}{!}
		{\begin{tabular}{ccc}
			\Xhline{3\arrayrulewidth}
                \rule{0pt}{10pt}
			 \bf Method   & \bf \# Queries & \bf AP \\ \hline
                \rule{0pt}{10pt}
                D-DETR (ArXiv'20) & 500 & 74.4 \\
                Sparse R-CNN (CVPR'21) & 500 & 76.0 \\                
                E2EDET (Sparse R-CNN) (CVPR'22) & 500 & 77.2 \\ 
                DDQ R-CNN (CVPR'23) & 300 & 83.5 \\
                DDQ DETR (CVPR'23) & 900 & 85.8 \\ \hline                
                \rule{0pt}{10pt}
                \bf Ours (Sparse R-CNN) & \bf 500 & \bf 77.7 \\
                \bf Ours (DDQ R-CNN) & \bf 300 & \bf 84.5 \\
                \bf Ours (DDQ DETR) & \bf 900 & \bf 86.4 \\
                \Xhline{3\arrayrulewidth}
		\end{tabular}
		}
	\end{center}
	\label{tab4} 
\end{table}

\begin{figure*}[t]
    \begin{center}
    \centerline{\includegraphics[width=0.95\linewidth]{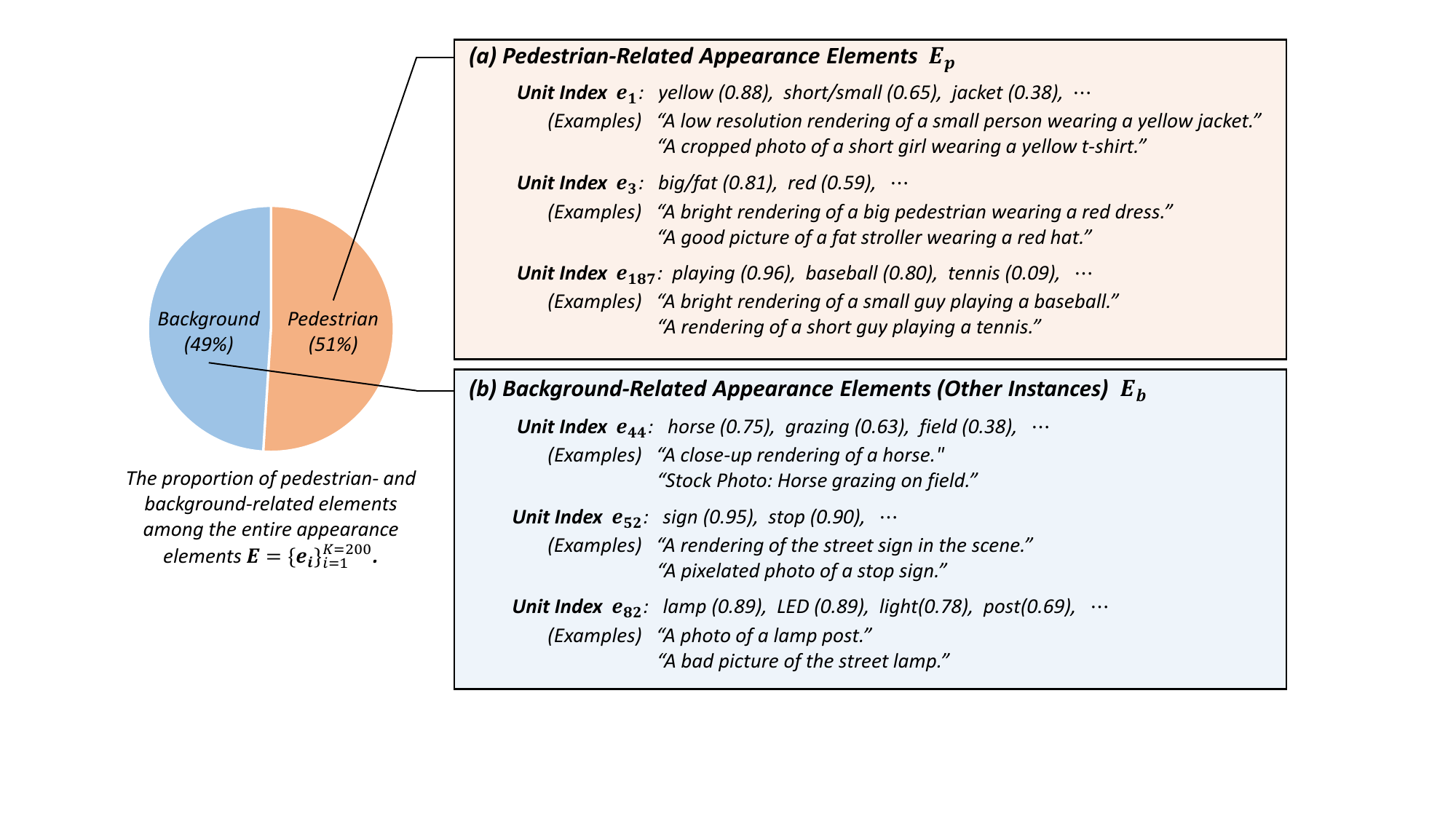}}
    \end{center}
    \caption{It shows the proportion of appearance elements for pedestrian and background. Among 200 appearance elements, the number of pedestrian-related appearance elements $\boldsymbol{E}_{p}$ is 102 (51\%), and the residual background-related elements $\boldsymbol{E}_{b}$ take 98 elements (49\%). (a) and (b) show the example elements and the descriptions mapped to each element for both $\boldsymbol{E}_{p}$ and $\boldsymbol{E}_{b}$. For example, ``\textit{A low resolution rendering of a small person wearing a yellow jacket.}'' and ``\textit{A cropped photo of a short girl wearing a yellow t-shirt.}'' belong to the first appearance element $\boldsymbol{e}_{1}$, one of the pedestrian-related elements.}
\label{fig4}
\end{figure*}

\subsection{Analysis on Language-Derived Appearance Elements}
We explore appearance elements $\boldsymbol{E}$. First, we find out which category each element represents, pedestrian-related appearance knowledge elements $\boldsymbol{E}_{p}$ and background-related elements $\boldsymbol{E}_{b}$. They are decided by inspecting whether the descriptions which are mapped to a specific element include a pedestrian-related name in \{\textit{class}\} or not. By doing so, we observe the proportion of $\boldsymbol{E}_{p}$ and $\boldsymbol{E}_{b}$. As described in Fig. \ref{fig4}, the diagram shows the proportion that $\boldsymbol{E}_{p}$ takes 51\% (102 elements) and $\boldsymbol{E}_{b}$ occupies 49\% (98 elements) among $K=200$ which is a default $K$. In more details, we explore the most representative contents that each element contains. Fig. \ref{fig4}(a) shows several examples of $\boldsymbol{E}_{p}$ and what they contain mostly. For example, the descriptions, such as ``\textit{A low resolution rendering of a small person wearing a yellow jacket.}'' and ``\textit{A cropped photo of a short girl wearing a yellow t-shirt.}'', usually belong to $\boldsymbol{e}_{1} \in \boldsymbol{E}_{p}$. It means that $\boldsymbol{e}_{1}$ mainly involves with appearance attributes representing `\textit{yellow color}', `\textit{short/small height}', and `\textit{jacket clothes} primarily. The numbers $0.88$ and $0.65$ next to each `\textit{yellow}' and `\textit{short/small}' mean that, among the descriptions mapped into $\boldsymbol{e}_{1}$, 88\% and 65\% of descriptions contain the words `\textit{yellow}' and `\textit{short/small}', respectively. For another pedestrian-related element $\boldsymbol{e}_{187}$, the descriptions regarding playing sports are aggregated frequently. For example, ``\textit{A bright rendering of a small guy playing a baseball.}'' and ``\textit{A rendering of a short guy playing a tennis.}'' belong to $\boldsymbol{e}_{187}$. On the other hand, several examples for $\boldsymbol{E}_{b}$ are described in Fig. \ref{fig4}(b). The words related to `\textit{horse}' and `\textit{grazing}' usually belong to $\boldsymbol{e}_{44} \in \boldsymbol{E}_{b}$ which is one of the background-related elements. It means that $\boldsymbol{e}_{44}$ can represent a horse category, one of background instances. The example descriptions are ``\textit{A close-up rendering of a horse.}'' and ``\textit{Stock Photo: Horse grazing on field.}''. Also, the descriptions, such as ``\textit{A photo of a lamp post.}'' and ``\textit{A bad picture of the street lamp.}'' are mapped into $\boldsymbol{e}_{82}$. It means that $\boldsymbol{e}_{82} \in \boldsymbol{E}_{b}$ contains a street light category primarily, which appears with pedestrians frequently in the street scene. As shown in the analysis, each element represents pedestrian- or background-related knowledge properly which could be helpful when being incorporated with visual cues in a pedestrian detection framework.

\subsection{Ablation Study}
We conduct ablation studies on \textit{CrowdHuman} \cite{crowdhuman} with respect to: 1) varying number of appearance knowledge elements $K$ and 2) the number of parameter increases, showing the effectiveness of our method that incorporates language-derived appearance elements.

\begin{figure}[t]
    \begin{center}
    \centerline{\includegraphics[width=0.99\linewidth]{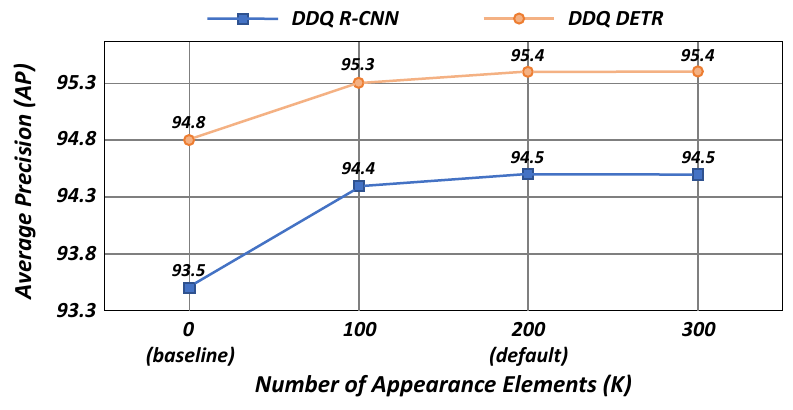}}
    \end{center}
    \caption{Ablation study with varying number of appearance elements $K$. We adopt average precision (AP) as evaluation metric. As shown in the figure, it shows consistent performance improvements while being insensitive to the number of $K$.}
\label{fig5}
\end{figure}

\subsubsection{Number of Appearance Elements $K$} Fig. \ref{fig5} shows the detection performances (\ie \ average precision, AP) with respect to varying $K$ of $0$, $100$, $200$, and $300$. Please note that $K=0$ means the baseline detection framework where appearance elements $\boldsymbol{E}$ are not incorporated. As shown in the figure, when we adopt $\boldsymbol{E}$ with small $K=100$, the detection performances improve on both pedestrian detection frameworks, DDQ R-CNN and DDQ DETR, compared to each baseline. Then when $K$ goes large, we can obtain consistent and stable performance improvements, and the detection performances become saturated. In other words, even though the number of appearance knowledge elements $K$ is a hyper-parameter that is required to be determined, it is not much sensitive to obtain performance improvement.

\subsubsection{Number of Parameters} We further analyze how much the number of parameters increases while incorporating language-derived appearance elements $\boldsymbol{E}$ with visual cues. Table \ref{tab5} shows the parameter increases with respect to the performance improvement of DDQ R-CNN and DDQ DETR, respectively. In this experiment, we further decompose the impact of $\boldsymbol{E}$ into $\boldsymbol{C}$ and $\boldsymbol{P}$ which is designed to make $\boldsymbol{C}$ to be more task-relevant. In other words, we also adopt and incorporate $\boldsymbol{C}$ directly with visual cues, instead of $\boldsymbol{E}=\boldsymbol{C} \oplus \boldsymbol{P}$. For DDQ R-CNN, $\boldsymbol{E}$ improves $1.0$AP with 5.6\% parameter increase only. Furthermore, we also conduct the experiment with DDQ DETR which is composed of 219M parameters. For DDQ DETR, $\boldsymbol{E}$ obtains a $0.6$AP performance gain with 3.6M parameter increase (1.6\%) only.

\begin{table}[t]
	\centering
    \caption{Table shows the effectiveness of language-derived appearance knowledge itself and the learnable prompts. We adopt average precision (AP) as evaluation metric. The proposed method brings performance gains with a small parameter increase for both detection frameworks.}
	\begin{center}
		\renewcommand{\tabcolsep}{5.0mm}
		\resizebox{0.99\linewidth}{!}
		{
			\begin{tabular}{ccc}
				\Xhline{3\arrayrulewidth}
                    \rule{0pt}{10pt}
			 	\bf Method & \bf \# Params & \bf AP \\ \hline
                    \rule{0pt}{10pt}
                    DDQ R-CNN (\textit{baseline}) & 63.5M   & 93.5  \\ \hdashline
                    \rule{0pt}{10pt}
                    \bf Ours ($\boldsymbol{C}$) & 67.1M (+5.6\%)    & 94.2  \\
                    \bf Ours ($\boldsymbol{U}=\boldsymbol{C} \oplus \boldsymbol{P}$) & 67.1M (+5.6\%) & 94.5  \\ \hline
                    \rule{0pt}{10pt}
                    DDQ DETR (\textit{baseline}) & 219M   & 94.8  \\ \hdashline
                    \rule{0pt}{10pt}
                    \bf Ours ($\boldsymbol{C}$) & 222.6M (+1.6\%)    & 95.2  \\
                    \bf Ours ($\boldsymbol{U}=\boldsymbol{C} \oplus \boldsymbol{P}$) & 222.6M (+1.6\%) & 95.4  \\ \Xhline{3\arrayrulewidth}
			\end{tabular}
		}
	\end{center}
	\label{tab5} 
\end{table}

\section{Discussions}
\subsection{Theoretical Basis}
The proposed method, which leverages language-derived appearance variation elements, is theoretically supported by the concept that external structured modalities, such as text, can effectively convey knowledge and assist in obtaining semantic-rich and fine-grained visual representations \cite{shen2022k, li2022grounded}. By integrating textual information into the visual understanding process, our method aligns with established theory in multimodal learning, where combining different sources of information can enhance overall comprehension. Furthermore, the usage of a learnable prompt for task-prompting is theoretically based on the understanding that there frequently exists a semantic gap between the knowledge encoded in a large-scale pretrained model and downstream tasks. This gap arises due to disparities in training objectives and data distributions \cite{petryk2022guiding, vt-clip, gondal2024domain}. By introducing a task-specific prompt, our method aims to bridge this semantic gap by guiding the appearance elements towards task-related information.

\subsection{Limitations}
The proposed method composes appearance description corpus illustrating diverse appearances of numerous objects. Our method has demonstrated the effectiveness consistently throughout experiments and analysis as introduced in the manuscript. However, it is worth noting that there could be instances where our method may mistakenly identify the figure in a portrait or poster as a pedestrian due to their similarity in appearance with real pedestrians. So, a current limitation of our method is that there are scenarios where additional contextual information regarding the surroundings of the pedestrians may be necessary for accurate detection. Therefore, one of the interesting future directions is not only to focus on providing the appearance itself but also to consider the surroundings of pedestrians for the better understanding of the nearby environment in the scene.

\section{Conclusion}
In this paper, we introduced a novel method to take advantage of an LLM in contextual appearance variation understanding and to leverage its knowledge into a pedestrian detection. We built a description corpus that contains abundant narratives for diverse appearances of pedestrians and others. We took them into an LLM and extracted appearance knowledge sets composed of the plentiful representations of appearance variations. Through task-prompting, we acquire appearance knowledge elements which are representative and become relevant with pedestrian detection. Finally, we provided language-derived appearance knowledge by incorporating the elements with visual cues in a pedestrian detector. With extensive experiments, we verified the effectiveness of our method showing state-of-the-art performances. Even though our method is designed for vision-language integration in pedestrian detection, we hope that the proposed method can provide useful insight to a wide range of research utilizing a variety of modalities.

\ifCLASSOPTIONcaptionsoff
  \newpage
\fi

\bibliographystyle{IEEEtran}
\bibliography{main}

\begin{thebibliography}{10}
\providecommand{\url}[1]{#1}
\csname url@samestyle\endcsname
\providecommand{\newblock}{\relax}
\providecommand{\bibinfo}[2]{#2}
\providecommand{\BIBentrySTDinterwordspacing}{\spaceskip=0pt\relax}
\providecommand{\BIBentryALTinterwordstretchfactor}{4}
\providecommand{\BIBentryALTinterwordspacing}{\spaceskip=\fontdimen2\font plus
\BIBentryALTinterwordstretchfactor\fontdimen3\font minus \fontdimen4\font\relax}
\providecommand{\BIBforeignlanguage}[2]{{%
\expandafter\ifx\csname l@#1\endcsname\relax
\typeout{** WARNING: IEEEtran.bst: No hyphenation pattern has been}%
\typeout{** loaded for the language `#1'. Using the pattern for}%
\typeout{** the default language instead.}%
\else
\language=\csname l@#1\endcsname
\fi
#2}}
\providecommand{\BIBdecl}{\relax}
\BIBdecl

\bibitem{context1}
E.~Kasneci, K.~Se{\ss}ler, S.~K{\"u}chemann, M.~Bannert, D.~Dementieva, F.~Fischer, U.~Gasser, G.~Groh, S.~G{\"u}nnemann, E.~H{\"u}llermeier \emph{et~al.}, ``Chatgpt for good? on opportunities and challenges of large language models for education,'' \emph{Learning and Individual Differences}, vol. 103, p. 102274, 2023.

\bibitem{gpt-3}
T.~Brown, B.~Mann, N.~Ryder, M.~Subbiah, J.~D. Kaplan, P.~Dhariwal, A.~Neelakantan, P.~Shyam, G.~Sastry, A.~Askell \emph{et~al.}, ``Language models are few-shot learners,'' \emph{Advances in Neural Information Processing Systems}, vol.~33, pp. 1877--1901, 2020.

\bibitem{ctrl}
N.~S. Keskar, B.~McCann, L.~R. Varshney, C.~Xiong, and R.~Socher, ``Ctrl: A conditional transformer language model for controllable generation,'' \emph{arXiv preprint arXiv:1909.05858}, 2019.

\bibitem{bert}
J.~Devlin, M.-W. Chang, K.~Lee, and K.~Toutanova, ``Bert: Pre-training of deep bidirectional transformers for language understanding,'' \emph{arXiv preprint arXiv:1810.04805}, 2018.

\bibitem{context2}
S.~Gururangan, A.~Marasovi{\'c}, S.~Swayamdipta, K.~Lo, I.~Beltagy, D.~Downey, and N.~A. Smith, ``Don't stop pretraining: Adapt language models to domains and tasks,'' \emph{arXiv preprint arXiv:2004.10964}, 2020.

\bibitem{csvt_lang1}
X.~Yang, F.~Lv, F.~Liu, and G.~Lin, ``Self-training vision language berts with a unified conditional model,'' \emph{IEEE Transactions on Circuits and Systems for Video Technology}, 2023.

\bibitem{csvt_lang2}
C.~Ma, Y.~Liu, J.~Deng, L.~Xie, W.~Dong, and C.~Xu, ``Understanding and mitigating overfitting in prompt tuning for vision-language models,'' \emph{IEEE Transactions on Circuits and Systems for Video Technology}, 2023.

\bibitem{vild}
X.~Gu, T.-Y. Lin, W.~Kuo, and Y.~Cui, ``Open-vocabulary object detection via vision and language knowledge distillation,'' \emph{arXiv preprint arXiv:2104.13921}, 2021.

\bibitem{labo}
Y.~Yang, A.~Panagopoulou, S.~Zhou, D.~Jin, C.~Callison-Burch, and M.~Yatskar, ``Language in a bottle: Language model guided concept bottlenecks for interpretable image classification,'' in \emph{Proceedings of the IEEE/CVF Conference on Computer Vision and Pattern Recognition}, 2023, pp. 19\,187--19\,197.

\bibitem{i2mvformer}
M.~F. Naeem, M.~G. Z.~A. Khan, Y.~Xian, M.~Z. Afzal, D.~Stricker, L.~Van~Gool, and F.~Tombari, ``I2mvformer: Large language model generated multi-view document supervision for zero-shot image classification,'' in \emph{Proceedings of the IEEE/CVF Conference on Computer Vision and Pattern Recognition}, 2023, pp. 15\,169--15\,179.

\bibitem{lavila}
Y.~Zhao, I.~Misra, P.~Kr{\"a}henb{\"u}hl, and R.~Girdhar, ``Learning video representations from large language models,'' in \emph{Proceedings of the IEEE/CVF Conference on Computer Vision and Pattern Recognition}, 2023, pp. 6586--6597.

\bibitem{visionllm}
W.~Wang, Z.~Chen, X.~Chen, J.~Wu, X.~Zhu, G.~Zeng, P.~Luo, T.~Lu, J.~Zhou, Y.~Qiao \emph{et~al.}, ``Visionllm: Large language model is also an open-ended decoder for vision-centric tasks,'' \emph{arXiv preprint arXiv:2305.11175}, 2023.

\bibitem{palm}
A.~Chowdhery, S.~Narang, J.~Devlin, M.~Bosma, G.~Mishra, A.~Roberts, P.~Barham, H.~W. Chung, C.~Sutton, S.~Gehrmann \emph{et~al.}, ``Palm: Scaling language modeling with pathways,'' \emph{arXiv preprint arXiv:2204.02311}, 2022.

\bibitem{shao-vqa}
Z.~Shao, Z.~Yu, M.~Wang, and J.~Yu, ``Prompting large language models with answer heuristics for knowledge-based visual question answering,'' in \emph{Proceedings of the IEEE/CVF Conference on Computer Vision and Pattern Recognition}, 2023, pp. 14\,974--14\,983.

\bibitem{t5}
C.~Raffel, N.~Shazeer, A.~Roberts, K.~Lee, S.~Narang, M.~Matena, Y.~Zhou, W.~Li, and P.~J. Liu, ``Exploring the limits of transfer learning with a unified text-to-text transformer,'' \emph{The Journal of Machine Learning Research}, vol.~21, no.~1, pp. 5485--5551, 2020.

\bibitem{csvt_ped1}
Y.~Jiao, H.~Yao, and C.~Xu, ``Pen: Pose-embedding network for pedestrian detection,'' \emph{IEEE Transactions on Circuits and Systems for Video Technology}, vol.~31, no.~3, pp. 1150--1162, 2020.

\bibitem{csvt_ped2}
J.~U. Kim, S.~Park, and Y.~M. Ro, ``Uncertainty-guided cross-modal learning for robust multispectral pedestrian detection,'' \emph{IEEE Transactions on Circuits and Systems for Video Technology}, vol.~32, no.~3, pp. 1510--1523, 2021.

\bibitem{csvt_ped3}
T.~Liu, K.-M. Lam, R.~Zhao, and G.~Qiu, ``Deep cross-modal representation learning and distillation for illumination-invariant pedestrian detection,'' \emph{IEEE Transactions on Circuits and Systems for Video Technology}, vol.~32, no.~1, pp. 315--329, 2021.

\bibitem{csvt_ped4}
J.~Li, Y.~Bi, S.~Wang, and Q.~Li, ``Cfrla-net: A context-aware feature representation learning anchor-free network for pedestrian detection,'' \emph{IEEE Transactions on Circuits and Systems for Video Technology}, 2023.

\bibitem{csvt_ped5}
C.~Lin, J.~Lu, and J.~Zhou, ``Multi-grained deep feature learning for robust pedestrian detection,'' \emph{IEEE Transactions on Circuits and Systems for Video Technology}, vol.~29, no.~12, pp. 3608--3621, 2018.

\bibitem{csvt_ped6}
X.~Wang, C.~Liang, C.~Chen, J.~Chen, Z.~Wang, Z.~Han, and C.~Xiao, ``S3d: scalable pedestrian detection via score scale surface discrimination,'' \emph{IEEE Transactions on Circuits and Systems for Video Technology}, vol.~30, no.~10, pp. 3332--3344, 2019.

\bibitem{crowdhuman}
S.~Shao, Z.~Zhao, B.~Li, T.~Xiao, G.~Yu, X.~Zhang, and J.~Sun, ``Crowdhuman: A benchmark for detecting human in a crowd,'' \emph{arXiv preprint arXiv:1805.00123}, 2018.

\bibitem{wider}
C.~C. Loy, D.~Lin, W.~Ouyang, Y.~Xiong, S.~Yang, Q.~Huang, D.~Zhou, W.~Xia, Q.~Li, P.~Luo \emph{et~al.}, ``Wider face and pedestrian challenge 2018: Methods and results,'' \emph{arXiv preprint arXiv:1902.06854}, 2019.

\bibitem{llm1}
H.~Touvron, T.~Lavril, G.~Izacard, X.~Martinet, M.-A. Lachaux, T.~Lacroix, B.~Rozi{\`e}re, N.~Goyal, E.~Hambro, F.~Azhar \emph{et~al.}, ``Llama: Open and efficient foundation language models,'' \emph{arXiv preprint arXiv:2302.13971}, 2023.

\bibitem{llm2}
H.~Touvron, L.~Martin, K.~Stone, P.~Albert, A.~Almahairi, Y.~Babaei, N.~Bashlykov, S.~Batra, P.~Bhargava, S.~Bhosale \emph{et~al.}, ``Llama 2: Open foundation and fine-tuned chat models,'' \emph{arXiv preprint arXiv:2307.09288}, 2023.

\bibitem{llm3}
M.~R. Costa-juss{\`a}, J.~Cross, O.~{\c{C}}elebi, M.~Elbayad, K.~Heafield, K.~Heffernan, E.~Kalbassi, J.~Lam, D.~Licht, J.~Maillard \emph{et~al.}, ``No language left behind: Scaling human-centered machine translation,'' \emph{arXiv preprint arXiv:2207.04672}, 2022.

\bibitem{llm4}
E.~Almazrouei, H.~Alobeidli, A.~Alshamsi, A.~Cappelli, R.~Cojocaru, M.~Debbah, E.~Goffinet, D.~Heslow, J.~Launay, Q.~Malartic \emph{et~al.}, ``Falcon-40b: an open large language model with state-of-the-art performance,'' \emph{Findings of the Association for Computational Linguistics: ACL}, vol. 2023, pp. 10\,755--10\,773, 2023.

\bibitem{llm5}
L.~Xue, N.~Constant, A.~Roberts, M.~Kale, R.~Al-Rfou, A.~Siddhant, A.~Barua, and C.~Raffel, ``mt5: A massively multilingual pre-trained text-to-text transformer,'' \emph{arXiv preprint arXiv:2010.11934}, 2020.

\bibitem{gpt}
A.~Radford, K.~Narasimhan, T.~Salimans, I.~Sutskever \emph{et~al.}, ``Improving language understanding by generative pre-training,'' \emph{OpenAI blog}, 2018.

\bibitem{clip}
A.~Radford, J.~W. Kim, C.~Hallacy, A.~Ramesh, G.~Goh, S.~Agarwal, G.~Sastry, A.~Askell, P.~Mishkin, J.~Clark \emph{et~al.}, ``Learning transferable visual models from natural language supervision,'' in \emph{International Conference on Machine Learning}.\hskip 1em plus 0.5em minus 0.4em\relax PMLR, 2021, pp. 8748--8763.

\bibitem{vilbert}
J.~Lu, D.~Batra, D.~Parikh, and S.~Lee, ``Vilbert: Pretraining task-agnostic visiolinguistic representations for vision-and-language tasks,'' \emph{Advances in Neural Information Processing Systems}, vol.~32, 2019.

\bibitem{csvt_vis_lang1}
L.~Wang, Z.~He, R.~Dang, H.~Chen, C.~Liu, and Q.~Chen, ``Res-sts: Referring expression speaker via self-training with scorer for goal-oriented vision-language navigation,'' \emph{IEEE Transactions on Circuits and Systems for Video Technology}, 2023.

\bibitem{csvt_vis_lang2}
M.~Gao, J.~Yang, J.~Han, K.~Lu, F.~Zheng, and G.~Montana, ``Decoupling multimodal transformers for referring video object segmentation,'' \emph{IEEE Transactions on Circuits and Systems for Video Technology}, 2023.

\bibitem{csvt_vis_lang3}
H.~Zhu, C.~Zhang, Y.~Wei, S.~Huang, and Y.~Zhao, ``Esa: External space attention aggregation for image-text retrieval,'' \emph{IEEE Transactions on Circuits and Systems for Video Technology}, 2023.

\bibitem{csvt_vis_lang4}
Y.~Zheng, B.~Zhong, Q.~Liang, G.~Li, R.~Ji, and X.~Li, ``Towards unified token learning for vision-language tracking,'' \emph{IEEE Transactions on Circuits and Systems for Video Technology}, 2023.

\bibitem{ped3}
S.~G. Santos, T.~I. Ren, G.~D. Cavalcanti, T.~I. Jyh, and J.~Sijbers, ``Pedestrian detection under progressive occlusion,'' in \emph{2013 IEEE International Conference on Systems, Man, and Cybernetics}.\hskip 1em plus 0.5em minus 0.4em\relax IEEE, 2013, pp. 4322--4327.

\bibitem{ped1}
C.~Lin, J.~Lu, G.~Wang, and J.~Zhou, ``Graininess-aware deep feature learning for pedestrian detection,'' in \emph{Proceedings of the European Conference on Computer Vision}, 2018, pp. 732--747.

\bibitem{ped4}
C.~Chi, S.~Zhang, J.~Xing, Z.~Lei, S.~Z. Li, and X.~Zou, ``Pedhunter: Occlusion robust pedestrian detector in crowded scenes,'' in \emph{Proceedings of the AAAI Conference on Artificial Intelligence}, vol.~34, no.~07, 2020, pp. 10\,639--10\,646.

\bibitem{ped5}
M.~Liu, C.~Zhu, J.~Wang, and X.-C. Yin, ``Adaptive pattern-parameter matching for robust pedestrian detection,'' in \emph{Proceedings of the AAAI Conference on Artificial Intelligence}, vol.~35, no.~3, 2021, pp. 2154--2162.

\bibitem{deng_aaai}
J.~Deng, D.~Fan, X.~Qiu, and F.~Zhou, ``Improving crowded object detection via copy-paste,'' in \emph{Proceedings of the AAAI Conference on Artificial Intelligence}, vol.~37, no.~1, 2023, pp. 497--505.

\bibitem{csvt_ped7}
K.~Kumar and R.~K. Mishra, ``A diagonally oriented novel feature extractor for pedestrian detection and its efficient hardware implementation,'' \emph{IEEE Transactions on Circuits and Systems for Video Technology}, vol.~32, no.~4, pp. 2035--2042, 2021.

\bibitem{csvt_ped8}
S.~Zhang, C.~Bauckhage, D.~A. Klein, and A.~B. Cremers, ``Exploring human vision driven features for pedestrian detection,'' \emph{IEEE Transactions on Circuits and Systems for Video Technology}, vol.~25, no.~10, pp. 1709--1720, 2015.

\bibitem{csvt_ped9}
M.~Bilal, A.~Khan, M.~U.~K. Khan, and C.-M. Kyung, ``A low-complexity pedestrian detection framework for smart video surveillance systems,'' \emph{IEEE Transactions on Circuits and Systems for Video Technology}, vol.~27, no.~10, pp. 2260--2273, 2016.

\bibitem{csvt_ped10}
W.~Ouyang, X.~Zeng, and X.~Wang, ``Partial occlusion handling in pedestrian detection with a deep model,'' \emph{IEEE Transactions on Circuits and Systems for Video Technology}, vol.~26, no.~11, pp. 2123--2137, 2015.

\bibitem{li2017adaptive}
X.~Li, M.~Ye, Y.~Liu, and C.~Zhu, ``Adaptive deep convolutional neural networks for scene-specific object detection,'' \emph{IEEE Transactions on Circuits and Systems for Video Technology}, vol.~29, no.~9, pp. 2538--2551, 2017.

\bibitem{ped2}
J.~Zhang, L.~Lin, J.~Zhu, Y.~Li, Y.-c. Chen, Y.~Hu, and S.~C. Hoi, ``Attribute-aware pedestrian detection in a crowd,'' \emph{IEEE Transactions on Multimedia}, vol.~23, pp. 3085--3097, 2020.

\bibitem{e2edet}
A.~Zheng, Y.~Zhang, X.~Zhang, X.~Qi, and J.~Sun, ``Progressive end-to-end object detection in crowded scenes,'' in \emph{Proceedings of the IEEE/CVF Conference on Computer Vision and Pattern Recognition}, 2022, pp. 857--866.

\bibitem{ddq}
S.~Zhang, X.~Wang, J.~Wang, J.~Pang, C.~Lyu, W.~Zhang, P.~Luo, and K.~Chen, ``Dense distinct query for end-to-end object detection,'' in \emph{Proceedings of the IEEE/CVF Conference on Computer Vision and Pattern Recognition}, 2023, pp. 7329--7338.

\bibitem{tang2023otp}
Y.~Tang, M.~Liu, B.~Li, Y.~Wang, and W.~Ouyang, ``Otp-nms: Towards optimal threshold prediction of nms for crowded pedestrian detection,'' \emph{IEEE Transactions on Image Processing}, 2023.

\bibitem{tu2018semantic}
Z.~Tu, W.~Xie, J.~Dauwels, B.~Li, and J.~Yuan, ``Semantic cues enhanced multimodality multistream cnn for action recognition,'' \emph{IEEE Transactions on Circuits and Systems for Video Technology}, vol.~29, no.~5, pp. 1423--1437, 2018.

\bibitem{cao2021handcrafted}
J.~Cao, Y.~Pang, J.~Xie, F.~S. Khan, and L.~Shao, ``From handcrafted to deep features for pedestrian detection: A survey,'' \emph{IEEE Transactions on Pattern Analysis and Machine Intelligence}, vol.~44, no.~9, pp. 4913--4934, 2021.

\bibitem{vlpd}
M.~Liu, J.~Jiang, C.~Zhu, and X.-C. Yin, ``Vlpd: Context-aware pedestrian detection via vision-language semantic self-supervision,'' in \emph{Proceedings of the IEEE/CVF Conference on Computer Vision and Pattern Recognition}, 2023, pp. 6662--6671.

\bibitem{wordnet}
G.~A. Miller, ``Wordnet: a lexical database for english,'' \emph{Communications of the ACM}, vol.~38, no.~11, pp. 39--41, 1995.

\bibitem{wordnet2}
C.~Fellbaum, ``Wordnet,'' in \emph{Theory and Applications of Ontology: Computer Applications}.\hskip 1em plus 0.5em minus 0.4em\relax Springer, 2010, pp. 231--243.

\bibitem{coco}
T.-Y. Lin, M.~Maire, S.~Belongie, J.~Hays, P.~Perona, D.~Ramanan, P.~Doll{\'a}r, and C.~L. Zitnick, ``Microsoft coco: Common objects in context,'' in \emph{Proceedings of the European Conference on Computer Vision}.\hskip 1em plus 0.5em minus 0.4em\relax Springer, 2014, pp. 740--755.

\bibitem{laion}
C.~Schuhmann, R.~Vencu, R.~Beaumont, R.~Kaczmarczyk, C.~Mullis, A.~Katta, T.~Coombes, J.~Jitsev, and A.~Komatsuzaki, ``Laion-400m: Open dataset of clip-filtered 400 million image-text pairs,'' \emph{arXiv preprint arXiv:2111.02114}, 2021.

\bibitem{adaprompt}
Y.~Chen, Y.~Liu, L.~Dong, S.~Wang, C.~Zhu, M.~Zeng, and Y.~Zhang, ``Adaprompt: Adaptive model training for prompt-based nlp,'' \emph{arXiv preprint arXiv:2202.04824}, 2022.

\bibitem{vt-clip}
L.~Qiu, R.~Zhang, Z.~Guo, Z.~Zeng, Y.~Li, and G.~Zhang, ``Vt-clip: Enhancing vision-language models with visual-guided texts,'' \emph{arXiv preprint arXiv:2112.02399}, 2021.

\bibitem{lxmert}
H.~Tan and M.~Bansal, ``Lxmert: Learning cross-modality encoder representations from transformers,'' \emph{arXiv preprint arXiv:1908.07490}, 2019.

\bibitem{att}
A.~Vaswani, N.~Shazeer, N.~Parmar, J.~Uszkoreit, L.~Jones, A.~N. Gomez, {\L}.~Kaiser, and I.~Polosukhin, ``Attention is all you need,'' \emph{Advances in Neural Information Processing Systems}, vol.~30, 2017.

\bibitem{nltk}
S.~Bird, E.~Klein, and E.~Loper, \emph{Natural language processing with Python: analyzing text with the natural language toolkit}.\hskip 1em plus 0.5em minus 0.4em\relax "O'Reilly Media, Inc.", 2009.

\bibitem{sentence-t5}
J.~Ni, G.~H. {\'A}brego, N.~Constant, J.~Ma, K.~B. Hall, D.~Cer, and Y.~Yang, ``Sentence-t5: Scalable sentence encoders from pre-trained text-to-text models,'' \emph{arXiv preprint arXiv:2108.08877}, 2021.

\bibitem{sparse}
P.~Sun, R.~Zhang, Y.~Jiang, T.~Kong, C.~Xu, W.~Zhan, M.~Tomizuka, L.~Li, Z.~Yuan, C.~Wang \emph{et~al.}, ``Sparse r-cnn: End-to-end object detection with learnable proposals,'' in \emph{Proceedings of the IEEE/CVF Conference on Computer Vision and Pattern Recognition}, 2021, pp. 14\,454--14\,463.

\bibitem{d-detr}
X.~Zhu, W.~Su, L.~Lu, B.~Li, X.~Wang, and J.~Dai, ``Deformable detr: Deformable transformers for end-to-end object detection,'' \emph{arXiv preprint arXiv:2010.04159}, 2020.

\bibitem{pmip}
J.~Wang, C.~Zhao, Z.~Huo, Y.~Qiao, and H.~Sima, ``High quality proposal feature generation for crowded pedestrian detection,'' \emph{Pattern Recognition}, vol. 128, p. 108605, 2022.

\bibitem{resnet}
K.~He, X.~Zhang, S.~Ren, and J.~Sun, ``Deep residual learning for image recognition,'' in \emph{Proceedings of the IEEE/CVF Conference on Computer Vision and Pattern Recognition}, 2016, pp. 770--778.

\bibitem{swin}
Z.~Liu, Y.~Lin, Y.~Cao, H.~Hu, Y.~Wei, Z.~Zhang, S.~Lin, and B.~Guo, ``Swin transformer: Hierarchical vision transformer using shifted windows,'' in \emph{Proceedings of the IEEE/CVF International Conference on Computer Vision}, 2021, pp. 10\,012--10\,022.

\bibitem{shen2022k}
S.~Shen, C.~Li, X.~Hu, Y.~Xie, J.~Yang, P.~Zhang, Z.~Gan, L.~Wang, L.~Yuan, C.~Liu \emph{et~al.}, ``K-lite: Learning transferable visual models with external knowledge,'' \emph{Advances in Neural Information Processing Systems}, vol.~35, pp. 15\,558--15\,573, 2022.

\bibitem{li2022grounded}
L.~H. Li, P.~Zhang, H.~Zhang, J.~Yang, C.~Li, Y.~Zhong, L.~Wang, L.~Yuan, L.~Zhang, J.-N. Hwang \emph{et~al.}, ``Grounded language-image pre-training,'' in \emph{Proceedings of the IEEE/CVF Conference on Computer Vision and Pattern Recognition}, 2022, pp. 10\,965--10\,975.

\bibitem{petryk2022guiding}
S.~Petryk, L.~Dunlap, K.~Nasseri, J.~Gonzalez, T.~Darrell, and A.~Rohrbach, ``On guiding visual attention with language specification,'' in \emph{Proceedings of the IEEE/CVF Conference on Computer Vision and Pattern Recognition}, 2022, pp. 18\,092--18\,102.

\bibitem{gondal2024domain}
M.~W. Gondal, J.~Gast, I.~A. Ruiz, R.~Droste, T.~Macri, S.~Kumar, and L.~Staudigl, ``Domain aligned clip for few-shot classification,'' in \emph{Proceedings of the IEEE/CVF Winter Conference on Applications of Computer Vision}, 2024, pp. 5721--5730.

\end{thebibliography}

\end{document}